\documentclass{article}
\usepackage{spconf,amsmath,graphicx}

\usepackage{enumitem}
\usepackage{multirow}
\setlist{nosep, leftmargin=14pt}
\usepackage{booktabs}
\usepackage{caption}
\usepackage{subcaption}
\usepackage{mwe} 
\usepackage{url}

\title{Anatomical Grounding Pre-training for Medical Phrase Grounding}
\name{Wenjun Zhang$^{\star}$ \qquad Shekhar S. Chandra$^{\star}$ \qquad Aaron Nicolson$^{\dagger}$}

\address{$^{\star}$University of Queensland\\$^{\dagger}$Australian e-Health Research Centre, CSIRO Health and Biosecurity, Brisbane, Australia}

\begin{document}
%
\maketitle

\begin{abstract}

Medical Phrase Grounding (MPG) maps radiological findings described in medical reports to specific regions in medical images. The primary obstacle hindering progress in MPG is the scarcity of annotated data available for training and validation. We propose anatomical grounding as an in-domain pre-training task that aligns anatomical terms with corresponding regions in medical images, leveraging large-scale datasets such as Chest ImaGenome. Our empirical evaluation on MS-CXR demonstrates that anatomical grounding pre-training significantly improves performance in both a zero-shot learning and fine-tuning setting, outperforming state-of-the-art MPG models. Our fine-tuned model achieved state-of-the-art performance on MS-CXR with an mIoU of 61.2, demonstrating the effectiveness of anatomical grounding pre-training for MPG.


\end{abstract}

\section{Introduction}
MPG involves mapping a descriptive phrase containing a radiological finding to a specific region in a medical image \cite{10.1007/978-3-031-43990-2_35}. An MPG model could be used to visually connect findings in a radiologist report---whether produced by radiologist or by automatic report generation model---to the corresponding regions in the images. Findings accompanied by their associated bounding boxes are easier to verify, enhancing the reliability of reported information \cite{bernstein_can_2023, 10204026, doi:10.1148/ryai.2020190043}.

MPG is a specialised application within the broader field of phrase grounding. State-of-the-art general-domain phrase grounding models are pre-trained on large-scale phrase-to-region datasets and demonstrate strong zero-shot learning and few-shot transferability on downstream localisation tasks \cite{9879567, 9710994, 10.1007/978-3-031-72970-6_3}. However, despite their success in general-domain tasks, these models struggle to generalise to MPG, especially in a zero-shot learning setting. One possible reason is the significant domain shift from general-domain to medical-domain data \cite{zhao-titov-2023-transferability}. Furthermore, large-scale pre-training is challenging in the medical domain due to the scarcity of annotated MPG datasets, with only a small public benchmark dataset available \cite{10.1007/978-3-031-20059-5_1}. 

\begin{figure}
    \centering
    \includegraphics[width=1\linewidth]{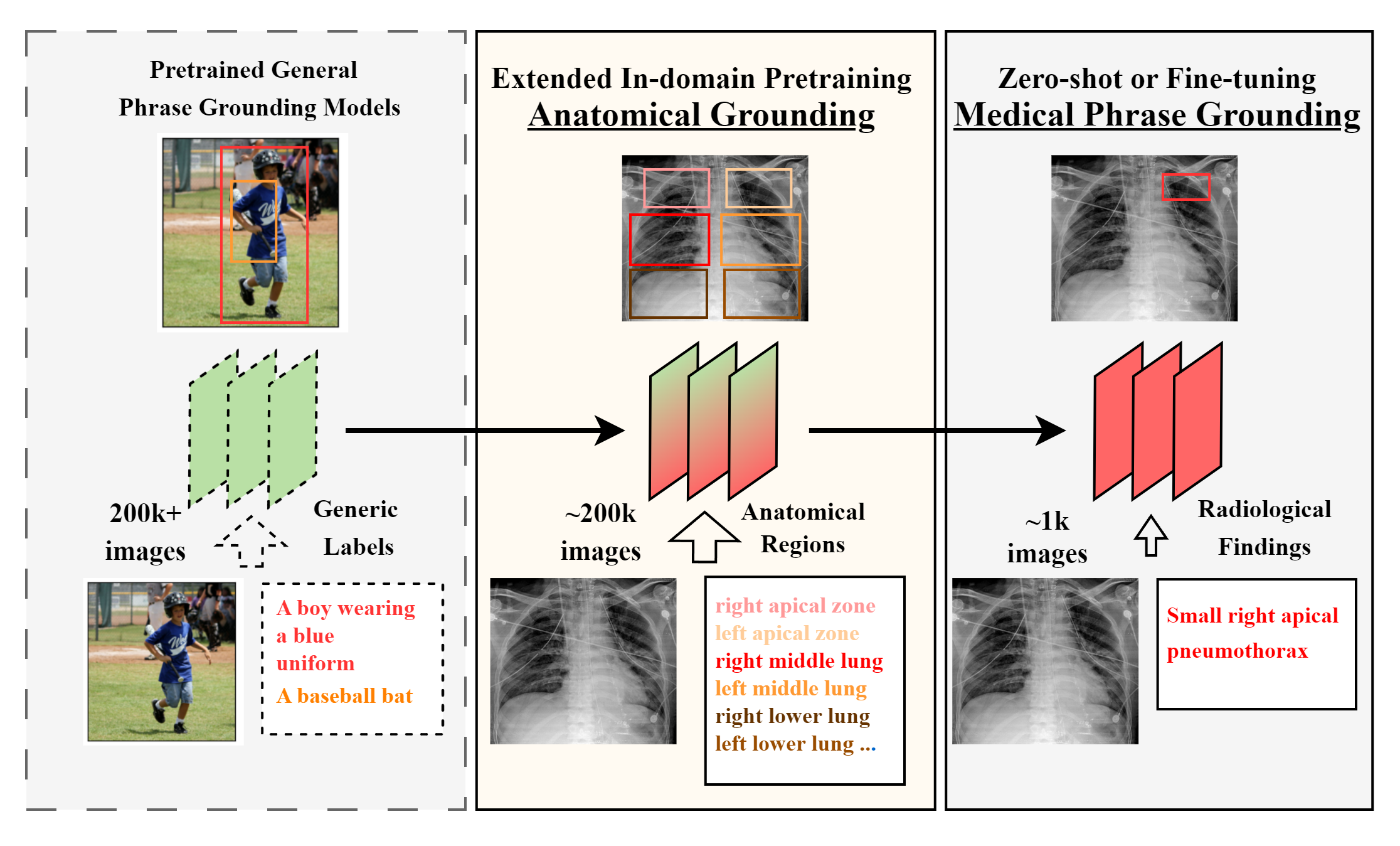}
    \caption{Anatomical grounding as an in-domain pre-training task for Medical Phrase Grounding (MPG).}
    \label{fig:concept}
\end{figure}

To overcome the challenges of limited MPG training data and the large domain gap between MPG and the general phrase grounding data, we propose to leverage anatomical grounding as an in-domain pre-training task for MPG, as demonstrated in Figure \ref{fig:concept} (middle). Anatomical grounding involves aligning text describing an anatomical region with the corresponding region within a medical image. This approach leverages the extensive anatomical text-to-region data available in datasets such as Chest ImaGenome \cite{wu2021chestimagenomedatasetclinical}, enabling effective fine-tuning or zero-shot learning for MPG tasks, where data is more limited \cite{ 10.1007/978-3-031-20059-5_1}. This pre-training step equips the model to recognise common anatomical landmarks, which radiologists frequently reference when describing findings in radiological reports. For instance, by learning to localise the \textit{right apical zone} with the Chest ImaGenome dataset, the model is more capable of localising findings such as a \textit{small right apical pneumothorax}.

We evaluated the effectiveness of anatomical grounding pre-training on MS-CXR, a MPG dataset, using two pre-trained general-domain phrase grounding models, TransVG \cite{9710016} and MDETR \cite{9710994}. We also evaluate it in both a zero-shot learning and a fine-tuning setting. Figure \ref{fig:concept} describes the training process; TransVG or MDETR is first pre-trained on anatomical grounding. They are then fine-tuned on MPG (if they are not evaluated in a zero-shot learning setting). Our empirical evaluation demonstrates that anatomical grounding pre-training significantly improves performance in a zero-shot learning setting, and significantly improves the performance of MDETR in a fine-tuning setting. We compare our anatomically grounded pre-trained models to state-of-the-art MPG models from the literature, and demonstrate that our models achieve an improvement in performance. The pre-trained models, and demo for this work are available at: \url{https://github.com/Claire1217/AGPT}.

\section{Related Work}
\subsection{General-domain Phrase Grounding}
Vision-language models pre-trained on large-scale image-text datasets, such as CLIP, have shown strong zero-shot learning and few-shot learning capabilities on global image understanding tasks \cite{pmlr-v139-radford21a}. GLIP extends this by pre-training on large-scale phrase grounding data \cite{9879567}. The learned representations demonstrate strong transferability to various local-level recognition tasks. Current pre-trained general-domain phrase grounding models are typically applied to two primary tasks: phrase localisation and referring expression comprehension. Phrase localisation focuses on identifying and locating multiple objects mentioned in a sentence. MDETR is a phrase localisation model, associating sub-phrases within a sentence with multiple object queries \cite{9710994}. In contrast, TransVG is a referring expression comprehension model---it detects a single object or region in an image for a whole sentence \cite{9710016}.

\subsection{Medical Phrase Grounding}
Due to the scarcity of annotated data, MPG has received limited attention in the literature. Boecking \textit{et al.} introduced MS-CXR, a phrase grounding chest X-ray benchmark dataset \cite{10.1007/978-3-031-20059-5_1}. Their objective with the dataset was to evaluate the grounding performance of their self-supervised biomedical vision-language model (BioViL). BioViL demonstrates strong zero-shot learning capabilities, given that it is not trained for MPG. Recently, Chen \textit{et al.} directly fine-tuned TransVG on a split of MS-CXR in order to directly learn MPG, forming MedRPG \cite{10.1007/978-3-031-43990-2_35}. Here, a bounding box supervised loss and a specific contrastive loss were leveraged. Unlike these models, we pre-train on large-scale anatomical grounding data using Chest ImaGenome, in order to provide in-domain pre-training.

\subsection{Anatomical Information in Medical Imaging}
Anatomical information has been effectively used in tasks like pathology detection and classification to improve accuracy and localisation. For example, the Anatomy-Driven Pathology Detection (ADPD) model \cite{muller_anatomy-driven_2023} used easy-to-annotate anatomical regions as proxies for pathologies, helping to locate disease locations without detailed pathology-specific bounding boxes. AnaXNet \cite{agu_anaxnet_2021} used anatomical relationships to improve classification by identifying the exact regions where findings occur. Despite these successes, no work has applied anatomical information to medical phrase grounding. 

\section{Methodology}\label{sec:methodology}
Our work addresses \textbf{medical phrase grounding} (MPG),  which involves mapping a descriptive phrase containing radiological finding to a specific
region in a medical image. This can be defined as learning a function  \( f: P \times I \rightarrow B \), where \( P \) represents the set of medical phrases, \( I \) represents the set of medical images, and \( B \) represents the set of bounding boxes. Given a phrase \( p \in P \) and an image \( i \in I \), the model predicts a bounding box \( b \in B \) such that $b = f(p, i)$. Our approach introduces a novel training framework for MPG, which involves extending the pre-training of general phrase grounding models with an anatomical grounding pre-training. 

Anatomical grounding involves predicting bounding boxes for anatomical structures using textual descriptions of their locations. The task can be formulated as 
\( f_{\text{anat}}: A \times I \rightarrow B \). Specifically, for each anatomical term \( a \in A \) and image \( i \in I \), the model predicts a bounding box \( b \in B \) such that $b = f_{\text{anat}}(a, i; \theta_{\text{gen}})$, 
where \( \theta_{\text{gen}} \) are the initial general-domain pre-trained weights. Through anatomical grounding pre-training, we refine the weights to create anatomy-specific parameters  \( \theta_{\text{anat}} \). 

To enhance generalisation and robustness, we leverage GPT-4 to generate four additional synonymous variations for each anatomical location in the Chest ImaGenome dataset. This aligns with clinical practice, where radiologists frequently use interchangeable terms to describe the same region. For example, ``left lung base” might also be referred to as ``left basal lung” or ``left lower lung base”. The detailed augmentation of anatomical regions is included in the aforementioned code repository. 

\section{Datasets} \label{sec:dataset}

\paragraph*{Chest ImaGenome \cite{wu2021chestimagenomedatasetclinical}}
We use the Chest ImaGenome dataset for anatomical grounding pre-training. Chest ImaGenome is a scene graph-structured dataset that includes $242\,072$ images. It contains $1\,256$ combinations of relational annotations between 29 anatomical structures in chest X-rays, with bounding box coordinates and additional attributes organised as a scene graph per image. In this study, we use the names and bounding box coordinates of these 29 anatomical structures, focusing specifically on frontal images. Examples of anatomical structures include ``left lung base", ``left lung apical zone", and ``right hilar structures".

\paragraph*{MS-CXR \cite{10.1007/978-3-031-20059-5_1}} 
We use the MS-CXR dataset for the MPG task. It contains $1\,162$ medical phrase-bounding box pairs across eight pathologies, such as \textit{cardiomegaly} and \textit{pleural effusion}. The findings are manually annotated and described by radiologists, ensuring precise alignment between medical phrases and bounding boxes. Example phrases include ``Large right-sided pneumothorax", and ``Small bilateral pleural effusions". The whole dataset was used for testing for the zero-shot learning setting with the general-domain pre-trained and anatomical pre-trained phrase grounding models, while the train-test-val split from \cite{10.1007/978-3-031-43990-2_35} was used for the fine-tuning setting. 

\section{Experiment Setup}
\paragraph*{Model}
Experiments were conducted with two models, TransVG and MDETR. For TransVG, ResNet-50 and ClinicalBERT were used as the visual and text encoders, respectively, whereas ResNet-101 and RoBERTa-base were used for MDETR. Here, MDETR functions on a sentence-level, mapping a medical phrase to one region in an image. This differs from its standard function, where it maps multiple sub-phrases from a sentence to multiple regions in the image. Full-model anatomical grounding pre-training of MDETR resulted in an unstable training process, likely due to its multi-object detection task. To address this, we applied Low-Rank Adaptation (LoRA) \cite{Hu2021LoRA:Models} during anatomical grounding pre-training. This likely stabilised training by limiting trainable parameters to low-rank layers, preventing drastic weight updates and reducing instability during adaptation.

\paragraph*{Pre-training and Fine-Tuning}
For anatomical grounding pre-training, we process mini-batches of eight images, each paired with five anatomical regions chosen from five synonymous terms, creating 40 anatomical text-region pairs per mini-batch. For MPG fine-tuning, both models were trained on the MS-CXR training set with a mini-batch size of 12. During fine-tuning, all of the weights of MDETR were trainable, including the LoRA weights. The AdamW optimiser with a learning rate of 1e-4 and 1e-5 was used for pre-training and fine-tuning, respectively \cite{DBLP:conf/iclr/LoshchilovH19}. Each model was trained for 1 epoch during pre-training and 90 epochs during fine-tuning. Images were resized and padded to a size of 640$\times$640. During training, the images were augmented with colour jitter and Gaussian noise.


\paragraph*{Evaluation}
We used mIoU and accuracy (Acc) as metrics. For accuracy, a predicted bounding box was considered true if the mIoU with the ground truth bounding box was larger than 0.5. We evaluate the anatomical grounding pre-trained MDETR and TransVG models on the MS-CXR dataset in both zero-shot learning and fine-tuning settings. The self-supervised pre-trained models GLoRIA \cite{9710099} and BioViL \cite{10.1007/978-3-031-20059-5_1} were used for comparison. In the fine-tuning setting, we further fine-tuned the anatomical grounding pre-trained MDETR and TransVG models on the training split of MS-CXR (described in Section \ref{sec:dataset}). These were compared to MDETR and TransVG without anatomical grounding pre-training and MedRPG \cite{10.1007/978-3-031-43990-2_35}. For zero-shot learning and fine-tuning, the epoch with the highest validation mIoU was selected for testing.

\section{Results \& Discussion}
\subsection{Effectiveness of Anatomical Grounding Pre-training}
The performance of anatomical grounding pre-training is demonstrated in Table \ref{tab:anat_comparison}. Applying MDETR and TransVG to MPG in a zero-shot learning setting produced low scores on both metrics, underscoring the limitations of general-domain phrase grounding models for MPG. However, pre-training with anatomical grounding led to a statistically significant improvement in both models’ performance across both metrics for zero-shot learning of MPG. These results demonstrate that anatomical grounding pre-training improves the models’ ability to generalise to MPG.

\begin{table}[ht]
\small
\centering
\caption{Performance of \textbf{anatomical grounding pre-training (AGPT)} on MS-CXR. Underlined indicates a stat. sig. difference to the model without anatomical grounding pre-training ($p < 0.05)$.}
\renewcommand{\arraystretch}{0.85}
\begin{tabular}{lcccc}
\toprule
\multirow{2}{*}{\textbf{Model}} & \multicolumn{2}{c}{\textbf{Zero-shot}} & \multicolumn{2}{c}{\textbf{Fine-tuning}} \\
\cmidrule(lr){2-3} \cmidrule(lr){4-5}
                                & \textbf{Acc}   & \textbf{mIoU}   & \textbf{Acc}    & \textbf{mIoU}   \\
\midrule
TransVG           & 1.2         & 10.3         & 68.9          & \textbf{59.4}          \\
\quad+ AGPT   & \textbf{\underline{39.8}}         & \textbf{\underline{40.7}}         & \textbf{70.7}          & 59.2 \\
\addlinespace 
MDETR              & 3.0         & 14.7         & 66.9          & 57.8         \\
\quad+ AGPT   &\textbf{ \underline{34.7}}         & \textbf{\underline{32.6}}         & \textbf{\underline{70.7} }         & \textbf{\underline{61.2}} \\
\bottomrule
\end{tabular}
\label{tab:anat_comparison}
\end{table}

When fine-tuning on the MS-CXR training set, anatomical grounding pre-training produced a statistically significant improvement across all metrics for MDETR. It also demonstrated an improvement with TransVG for Acc. This indicates that anatomical grounding pre-training is effective for MPG fine-tuning, particularly for certain types of models.

\begin{figure}[t]
    \centering
    \includegraphics[width=1\linewidth]{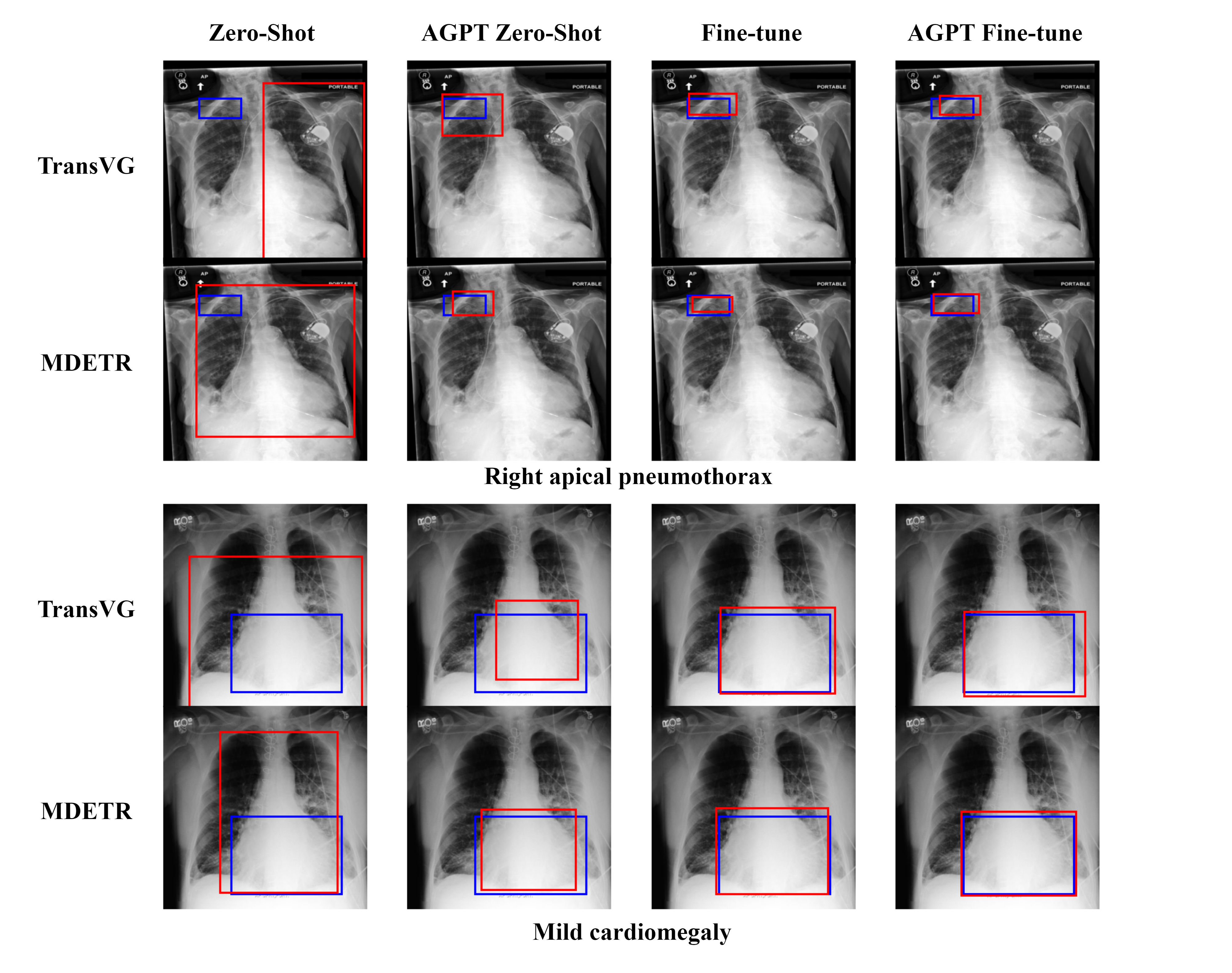}
    \caption{MPG with and without \textbf{anatomical grounding pre-training (AGPT)}. The top example contains the anatomical region within the text, whereas the bottom example does not. Blue and red boxes indicate the ground-truth and predicted bounding boxes, respectively.}
    \label{fig:viz}
\end{figure}

In Figure \ref{fig:viz}, we illustrate the models performing MPG in zero-shot learning settings on two examples: ``right apical pneumothorax" and ``mild cardiomegaly". Without anatomical grounding pre-training, both TransVG and MDETR fail to ground the phrases accurately. However, with anatomy pre-training, both models are able to ground the text to the correct anatomical region---the right apical zone for pneumothorax and the heart for cardiomegaly. Fine-tuning offers a further improvement in the grounding accuracy.

\begin{table*}[t]
\small
\centering
\caption{A comparison of \textbf{anatomical grounding pre-training (AGPT)} with other models in the literature in both zero-shot learning and fine-tuning settings with \textbf{mIoU} as the metric. $\dagger$ indicates scores sourced from the BioViL paper \cite{10.1007/978-3-031-20059-5_1}.}
\label{table:combined_iou_scores}
\renewcommand{\arraystretch}{0.85}
\begin{tabular}{l c c c c c c c c c c}
\toprule
\textbf{Model} & \textbf{Supervision} & \textbf{Cardio.} & \textbf{Opacity} & \textbf{Edema} & \textbf{Consol.} & \textbf{Pneu.} & \textbf{Atelect.} & \textbf{Pneumo.} & \textbf{Pl. Eff.} & \textbf{Avg} \\
\midrule
\multicolumn{11}{c}{\textbf{Zero-shot learning}} \\ 
\midrule
\textbf{GLoRIA \cite{9710099}}$^\dagger$ & Self-super. & 27.3 & 19.8 & 25.1 & 32.4 & 24.6 & 26.1 & 10.0 & 25.4 & 24.6 \\
\textbf{BioViL \cite{10.1007/978-3-031-20059-5_1}}$^\dagger$ & Self-super. & 37.5 & 20.9 & \textbf{27.5} & \textbf{34.6} & 31.5 & 30.2 & 13.5 & \textbf{31.5} & 28.4 \\
\cmidrule(lr){1-11}
\textbf{MDETR + AGPT} & Box-super. & 61.3 & 6.0 & 8.7 & 18.5 & 18.8 & 8.2 & 16.1 & 14.6 & 32.6 \\
\textbf{TransVG + AGPT} & Box-super. & \textbf{61.5} & \textbf{23.0} & 14.5 & 33.0 & \textbf{31.9} & \textbf{39.3} & \textbf{26.9} & 21.1 & \textbf{40.7} \\
\midrule
\multicolumn{11}{c}{\textbf{Fine-tuning}} \\ 
\midrule
\textbf{MedRPG \cite{10.1007/978-3-031-43990-2_35}} & Box-super. & 80.5 & 39.3 & \textbf{51.7} & 49.1 & 46.4 & \textbf{48.8} & 38.5 & 52.8 & 59.6 \\
\textbf{MDETR \cite{9710994}} & Box-super. & 79.6 & 43.1 & 45.5 & 45.8 & 40.1 & 36.0 & 39.1 & 50.5 & 57.8 \\
\textbf{TransVG \cite{9710016}} & Box-super. & 80.6 & \textbf{46.8} & 35.6 & 42.7 & \textbf{48.5} & 42.8 & 38.3 & 49.5 & 59.4 \\
\cmidrule(lr){1-11}
\textbf{MDETR + AGPT} & Box-super. & \textbf{81.2} & 45.1 & 25.2 & \textbf{56.3} & 38.9 & 47.4 & \textbf{43.1} & \textbf{57.2} & \textbf{61.2} \\
\textbf{TransVG + AGPT} & Box-super. & 79.1 & 37.6 & 43.0 & 45.4 & 45.9 & 47.7 & 41.9 & 54.1 & 59.2 \\
\bottomrule
\end{tabular}
\end{table*}
 
\subsection{Comparison to other MPG models}
First, we compare our anatomical grounding pre-trained MDETR and TransVG models (MDETR + AGPT and TransVG + AGPT, respectively) in a zero-shot learning setting, as shown in Table \ref{table:combined_iou_scores}. We compare these to two self-supervised models, GLoRIA \cite{9710099} and BioViL \cite{10.1007/978-3-031-20059-5_1}. Both MDETR + AGPT and TransVG + AGPT outperformed GLoRIA and BioViL. This indicates that anatomical grounding pre-training is more effective for zero-shot learning MPG than the self-supervised learning strategies of GLoRIA and BioViL. Furthermore, our fine-tuned MDETR + AGPT model attained an mIoU improvement of 1.6 over the current state-of-the-art model, MedRPG \cite{10.1007/978-3-031-43990-2_35}. 


\subsection{Effectiveness of Synonymous Anatomical Term Augmentation}
We conducted ablation studies to evaluate the impact of adding synonymous variations of the anatomical locations, as described in Section \ref{sec:methodology}. The results show that synonymous augmentation improved the scores for both TransVG and MDETR, with a stronger effect observed in TransVG. Notably, anatomical grounding pre-training with synonymous augmentation led to a 15.6\% improvement in zero-shot learning accuracy. This provides the model with a broader range of terms for the same anatomical location. This allows the model to better generalise to new phrases in a zero-shot learning setting.
\begin{table}[h!]
\small
\centering
\caption{Improvement in performance with when using synonymous variations of the anatomical locations. Underlined indicates a stat. sig. difference to the model without synonymous variations ($p < 0.05)$.}
\label{table:syn_augmentation_effect}
\renewcommand{\arraystretch}{0.85}
\begin{tabular}{lcccc}
\toprule
\multirow{2}{*}{\textbf{Model}}  & \multicolumn{2}{c}{\textbf{Zero-shot}} & \multicolumn{2}{c}{\textbf{Fine-tuning}} \\
\cmidrule(lr){2-3} \cmidrule(lr){4-5}
 & \textbf{Acc} & \textbf{mIoU} & \textbf{Acc} & \textbf{mIoU} \\
\midrule
TransVG   & \underline{+15.6} & \underline{+13.9} & \underline{+5.4} & \underline{+2.6} \\
MDETR     & \underline{+1.8}  & \underline{+1.2}   & +2.4 & +0.9 \\
\bottomrule
\end{tabular}
\end{table}

\vspace{-10pt}
\section{Conclusion}
In this paper, we introduced anatomical grounding pre-training to address the challenges of MPG, a task constrained by limited in-domain data and significant domain shifts from general-domain pre-trained models. Our methodology involved pre-training phrase grounding models on anatomical text-region pairs using the Chest ImaGenome dataset, followed by MPG-specific fine-tuning on the MS-CXR dataset. Empirical results demonstrated that anatomical grounding pre-training significantly improved zero-shot learning and fine-tuning performance on MPG, surpassing existing self-supervised and state-of-the-art MPG models. Additionally, our augmentation with synonymous anatomical terms further enhanced generalisability. This work demonstrates that leveraging anatomical grounding pre-training is an effective solution to the challenge of limited MPG data.

\section{Compliance with Ethical Standards}  
\vspace{-2mm} 
\noindent This study used public data from MIMIC-CXR (under PhysioNet’s credentialed license). Ethical approval was not required as confirmed by the license attached with the open access data.
\vspace{-2mm}

\section*{Acknowledgments}
\vspace{-2mm} 
\noindent No funding was received. The authors declare no competing interests.  

\vspace{-5mm} 

\bibliographystyle{IEEEtran}
\bibliography{isbi/ISBI_2024_template-master/references}
\end{document}